\documentclass[manuscript,screen,acmlarge]{acmart}

\usepackage{soul}
\usepackage{censor, caption}
\usepackage{subcaption}

\AtBeginDocument{%
  \providecommand\BibTeX{{%
    \normalfont B\kern-0.5em{\scshape i\kern-0.25em b}\kern-0.8em\TeX}}}

\copyrightyear{2023}
\acmYear{2023}
\acmDOI{}

\begin{document}

\title{Neuro-Symbolic Approaches for Context-Aware Human Activity Recognition}

\author{Luca Arrotta}
\email{luca.arrotta@unimi.it}
\orcid{0000-0001-5207-566X}
\author{Gabriele Civitarese}
\email{gabriele.civitarese@unimi.it}
\orcid{0000-0002-8247-2524}
\author{Claudio Bettini}
\email{claudio.bettini@unimi.it}
\orcid{0000-0002-1727-7650}
\affiliation{%
  \institution{University of Milan}
  \streetaddress{Via Celoria, 18}
  \city{Milan}
  \country{Italy}
}

\renewcommand{\shortauthors}{Arrotta, et al.}

\begin{abstract}

Deep Learning models are a standard solution for sensor-based Human Activity Recognition (HAR), but their deployment is often limited by labeled data scarcity and model's opacity. Neuro-Symbolic AI (NeSy) provides an interesting research direction to mitigate these issues by infusing knowledge about context information into HAR deep learning classifiers.
However, existing NeSy methods for context-aware HAR require computationally expensive symbolic reasoners during classification, making them less suitable for deployment on resource-constrained devices (e.g., mobile devices). Additionally, NeSy approaches for context-aware HAR have never been evaluated on in-the-wild datasets, and their generalization capabilities in real-world scenarios are questionable.
In this work, we propose a novel approach based on a semantic loss function that infuses knowledge constraints in the HAR model during the training phase, avoiding symbolic reasoning during classification. Our results on scripted and in-the-wild datasets show the impact of different semantic loss functions in outperforming a purely data-driven model. We also compare our solution with existing NeSy methods and analyze each approach's strengths and weaknesses. Our semantic loss remains the only NeSy solution that can be deployed as a single DNN without the need for symbolic reasoning modules, reaching recognition rates close (and better in some cases) to existing approaches.
\end{abstract}




%\received{15 February 2023}

\maketitle

\section{Introduction}
The sensor-based Human Activity Recognition (HAR) research area is dominated by solutions based on purely data-driven Deep Learning (DL) models~\cite{wang2019deep,chen2021deep}.
While DL-based solutions are very effective, they still have some open research issues that limit their deployment in real-world scenarios. Among the major problems, there are labeled data scarcity~\cite{abdallah2018activity} and the lack of transparency of the activity models~\cite{atzmueller2018explicative}.

In the literature, purely knowledge-based approaches have been proposed to tackle both problems~\cite{gayathri2017probabilistic}. Symbolic methods rely on domain knowledge (e.g., based on common-sense knowledge) to model constraints between sensor events and activities. The sensor data stream is then matched with symbolic rules to identify the most likely activities according to knowledge. Purely knowledge-based methods have two advantages: 1) they do not require labeled data, and 2) they are based on human-readable formalisms that make them interpretable and transparent.
However, these approaches are too rigid since it is unlikely that logic constraints can cover all the possible patterns related to activity execution. Moreover, they are not suitable for sensors that generate continuous values (e.g., accelerometer) since raw data can not be mapped to a clear semantic.

In the general machine learning community, Neuro-Symbolic AI (NeSy) methods are emerging to combine the strengths of data-driven and knowledge-based methods~\cite{hitzler2022neuro}. The idea of NeSy methods is to enhance DL models through domain knowledge. The potential advantages of NeSy are many. First, it may significantly improve the recognition rate by driving the classification with domain constraints. This may be especially true when only a limited amount of labeled data is available; hence, those constraints can not be learned directly from data. For the same reason, the use of domain knowledge can potentially improve the classification of those cases out of the training set distribution samples. %Moreover, domain knowledge may speed up the training process. Last but not least, 
Moreover, DL models enhanced through domain knowledge have the potential of being more interpretable and transparent, since their decisions are also influenced by the knowledge model~\cite{li2022interpretable}.

This work focuses on sensor-based HAR on mobile/wearable devices (e.g., smartphones, smartwatches). While the majority of existing works in this field only focus on inertial sensors, we also consider high-level context data (e.g., semantic position, weather) as also proposed by a few research groups~\cite{riboni2011cosar,cao2018gchar,asim2020context}.
This research area is usually referred to as \textit{Context-Aware Human Activity Recognition}.
%However, only a few works proposed to infuse knowledge in the DL models.
In the literature, a few NeSy approaches for context-aware HAR have been proposed by applying knowledge-based reasoning on high-level context data~\cite{arrotta2022knowledge}. To the best of our knowledge, such approaches have never been evaluated on public in-the-wild datasets, but only on small datasets acquired in a scripted fashion.

Moreover, the existing approaches in the literature involve symbolic reasoning during both training and classification. In real-world deployments, where the DL model is deployed on resource-constrained devices (e.g., mobile/wearable devices), the adoption of symbolic reasoning during classification is not desirable since it is too computationally demanding. Empirical experiments in the literature show that running symbolic reasoning on mobile devices is up to $150$ times slower than on machines with higher resources (e.g., servers)~\cite{bobed2015semantic}. In the HAR domain, the work in~\cite{bettini2020caviar} reports that context-aware ontological reasoning on mobile devices takes on average $1.3$ seconds for each data sample. % of $4$ second. 
Since samples are collected with high periodicity (e.g., a few seconds), such approaches may be inefficient in terms of computational resources. 

In this work, we propose a novel NeSy method for Context-Aware HAR on mobile devices. Our approach is based on a custom loss function that combines a standard classification loss and a novel semantic loss function based on symbolic reasoning. Our semantic loss drives the activity model to classify activities considering both raw sensor data patterns and high-level knowledge constraints. Indeed, after the training phase, the classifier internally encodes such constraints, that are exploited to classify activities without requiring symbolic reasoning.

Our experimental evaluation on scripted and in-the-wild datasets show that our method based on a semantic loss outperforms in terms of recognition rate a classic $DL$ approach based on a standard classification loss. We also compared our approach with two alternative NeSy strategies that use symbolic reasoning during classification, showing that our semantic loss often reaches recognition rates close (and sometimes better) to such state-of-the-art methods, while avoiding the significant cost of performing symbolic reasoning during inference. Hence, we believe that our semantic loss reaches a good trade-off between efficiency and recognition rate.

To summarize, our contributions are the following:
\begin{itemize}
    \item We formalize the NeSy Context-Aware HAR research problem and reformulate existing solutions using our notation.
    \item We propose a novel NeSy solution for Context-Aware HAR, based on a semantic loss function that does not require symbolic reasoning after training.
    \item We performed an extensive evaluation on scripted and in-the-wild datasets, comparing our solution with two existing NeSy methods that require symbolic reasoning during classification. Our results show that, especially considering in-the-wild settings, our semantic loss reaches recognition rates that are often close (and sometimes better) than the ones of the other approaches.
\end{itemize}

\section{Related work}
\label{sec:related}
Most of the works proposed in the literature for sensor-based HAR on mobile/wearable devices rely on supervised Deep Learning (DL) methods \cite{wang2019deep, chen2021deep}. 
The combination of inertial and high-level context data has the potential to significantly improve the recognition rate compared to considering only inertial sensors as proposed by the majority of the works~\cite{saguna2013complex}.

Despite their success, existing DL solutions require a large amount of labeled data during the learning process. Unfortunately, the annotation process is error-prone, expensive, time-consuming, and tedious, especially considering large amounts of data. Moreover, the inner mechanisms of deep learning classifiers are opaque, thus not allowing humans to understand the rationale behind each model's prediction.

To mitigate the data scarcity problem, the HAR research community investigated transfer learning, unsupervised learning, and semi-supervised learning approaches~\cite{chen2021deep}. Transfer learning methods usually take advantage of models trained on a source domain with a significant amount of labeled data. Such pre-trained models are then fine-tuned in a target domain using a small amount of labeled samples~\cite{sanabria2021unsupervised, soleimani2021cross}.
On the other hand, semi-supervised approaches for HAR rely on small labeled datasets to initialize the model, which is then incrementally updated by leveraging the unlabeled data stream~\cite{abdallah2018activity}. Semi-supervised methods for HAR include self-learning, co-learning, active learning, and label propagation.
Finally, unsupervised learning strategies can be used to derive activity clusters from the available large amounts of unlabeled data~\cite{kwon2014unsupervised, jain2022collossl,hiremath2022bootstrapping}. 
%However, unsupervised learning methods still require strategies to annotate clusters using labeled data.

In general, the major drawback of the above-mentioned solutions is that they are not conceived to be interpretable. Hence, they can be considered black boxes.
In the literature, the majority of interpretable models for HAR are not based on deep learning but on less effective inherently interpretable models~\cite{atzmueller2018explicative,guesgen2020using}. While a few works consider deep models~\cite{arrotta2022dexar}, they do not take into account data scarcity.

Neuro-Symbolic AI (NeSy) integrates neural and symbolic AI architectures to combine their abilities to perform learning and knowledge-based reasoning~\cite{kamruzzaman2021neuro}. This combination improves the capability of the deep learning classifier to learn from smaller amounts of training data, and, at the same time, it also increases its interpretability~\cite{gaur2021semantics}.
While most NeSy methods have been proposed for computer vision and NLP applications, only a few NeSy methods exist for HAR.
%~\cite{van2020indoor,azkune2018scalable}.
Considering HAR in smart-home environments, the domain knowledge can be used to derive an initial activity model that is subsequently adapted to the user's habits through data-driven strategies~\cite{sukor2019hybrid}. In~\cite{azkune2018scalable}, unsupervised methods are used to extract frequent patterns from unlabeled data. These patterns are then associated with the corresponding activities through domain knowledge. However, while these approaches are effective on smart-home environmental sensors, they cannot be applied to the inertial sensors data provided by mobile and wearable devices, which are the focus of this work. 

Considering context-aware HAR on mobile devices, \cite{bettini2020caviar} proposed to use domain knowledge on high-level context data to refine the predictions of an activity classifier trained on inertial sensors data. Finally, a recent work proposes the infusion of domain knowledge on context data into the deep learning classifier during both the training and inference phases~\cite{arrotta2022knowledge}. However, these approaches rely on ontological reasoning during classification, which may be critical for the deployment of resource-constrained mobile devices. More details about existing NeSy context-aware HAR approaches will be presented in Section~\ref{sec:revisiting}.

To the best of our knowledge, this is the first work that proposes a NeSy solution for HAR with the following characteristics: a) it infuses knowledge directly inside the DNN activity model during training, and b) it does not require symbolic reasoning during classification.

\section{Preliminaries}
\label{sec:problem}
In this section, we formalize context-aware HAR and we formulate the NeSy Context-Aware HAR problem. Moreover, we take advantage of this formalization to re-formulate existing NeSy strategies for Context-Aware HAR.

\subsection{Context-Aware Human Activity Recognition}
\label{sec:context-aware}

Let $D_{u}$ be the dataset of raw sensor data collected from the mobile devices (e.g., smartphone, smartwatch) of a user $u$.
%let $D^\star = \bigcup_{u \in U}D_{u}$ be the union of the datasets of the users $U$. 
Given a set of users $U=\{u_1, \dots, u_n\}$, 
let $D^\star = \{D_{u_1},\dots,D_{u_n}\}$ be the set of the datasets from all the users.
Let $A= \{a_1, \dots, a_k \}$ be the set of considered activities.
The dataset $D^\star$ is associated with a set of annotations $L$ that describes the activities performed by each user $u$. Each annotation $l \in L$ is a tuple $l = \langle u, a, t_s, t_e\rangle$ where $a$ is a label identifying the activity actually performed by $u$ during the time interval $[t_s, t_e]$. 
Each user dataset $D_{u}$ is partitioned in a set of non-overlapping fixed-length windows $W_u = \{w_1,\dots,w_q\}$ with each window including $z$ seconds of consecutive raw sensor data of $D_u$. 

In this work, we use the notion of \emph{context} as a specific high-level situation that occurs in the environment surrounding and including the user while sensor data are being acquired (e.g., \textit{it is raining}, \textit{location is a park}, \textit{current speed is high}). Let $C = \langle C_1, \dots, C_p \rangle$ be a set of possible contexts that are meaningful for the application domain.

Considering the recognition of physical activities (e.g., walking, sitting on transport), each window $w$ can be partitioned into $w^I$ and $w^C$. $w^I$ are sensor data that cannot be used to derive high-level contexts (e.g., data from inertial sensors). On the other hand, $w^C$ are those raw sensor data that can be used to derive high-level contexts in $C$ through reasoning and/or abstraction. More specifically, given a window $w=\langle w^C,w^I \rangle$, let $ca(w^C)$ be a \textit{Context Aggregation} function that derives all the contexts $C^w \subset C$ that are true during $w$ based on $w^C$. This function can rely on simple rules, available services, or context-aware middlewares~\cite{henricksen2005middleware}. For instance, the geographical coordinates provided by the location service of the user's smartphone can be used to derive her semantic location (e.g., at home, in a public park) by querying a dedicated web service.
%$w^C$ are those raw sensor data that can be used to derive high-level contexts in $C$ (e.g., geographical position). \hl{for example by interacting with dedicated web services}. On the other hand, $w^I$ are sensor data that cannot directly provide high-level contexts (e.g., data from inertial sensors).
%Given a window $w=\langle w^C,w^I \rangle$, let $ca(w^C)$ be a \textit{Context Aggregation} function that outputs all the contexts $C^w \subset C$ that are true during $w$. This function can rely on simple rules, available services, or context-aware middlewares~\cite{henricksen2005middleware}.
%For instance, GPS data can be used to obtain the semantic positions of the user (e.g., at home, in a public park) by querying a dedicated web service.

%\newtheorem{definition}{Definition}

\begin{definition}[Context-aware HAR]
Given a dataset $D^\star$ and the annotations set $L$, the problem of \textit{context-aware Human Activity Recognition} is to provide to an unseen tuple $\langle w^I, C^w \rangle$, derived from a sensor data window $w$ from user $u$, the probability distribution $P = \langle p_1, \dots, p_k \rangle$, where $p_i$ is the probability that $u$ performed the activity $a_i$ in contexts $C^w$, with $\sum_{i=1}^{k} p_i = 1$.
\end{definition}

\subsection{Neuro-Symbolic Context-Aware HAR}
\label{sec:neuro-symbolic}

%Given a deep neural network model $DNN$, the dataset $D^\star$, and the annotations set $L$, the goal of the training phase is to learn the weights of $DNN$ such that it can provide to an unseen tuple $\langle w^I, C^w \rangle$, derived from a sensor data window $w$ from user $u$, the probability distribution $P = \langle p_1, \dots, p_k \rangle$, where $p_i$ is the probability that $u$ performed the activity $a_i$, with $\sum_{i=1}^{k} p_i = 1$.

The \textit{context-aware HAR} problem could be tackled by using purely data-driven models where context data are simply used as input.
%Neuro-Symbolic approaches, that are based on a combination of deep learning and knowledge-based methods.
However, based on a set of contexts $C$, it is possible to build a knowledge model $K$ that encodes relationships between the activities in $A$ and the contexts in $C$. For instance, according to common-sense knowledge, the activity \textit{lying} is less likely performed in outdoor settings, like a public park, when it is raining. Note that $K$ can be built in several different ways: by domain experts using common-sense knowledge on HAR, re-using existing knowledge bases (e.g., ontologies), or considering semi-automatic approaches in charge of extracting knowledge from external sources (e.g., text, images, and videos from the web). 
% This knowledge should not necessarily come from the knowledge engineer and domain experts but it may be extracted semi-automatically in several ways, including: Proposing a survey to a large number of users; Scraping information about context and activities from the Web

Given a knowledge model $K$ and a set of contexts $C^w$, let $SR(K,C^w)$ be a  \textsc{symbolic reasoning} function that outputs the set of activities $A^\star$ that are consistent with $C^w$ according to the constraints in $K$.

\begin{figure}[h!]
\centering
\includegraphics[width=0.5\linewidth]{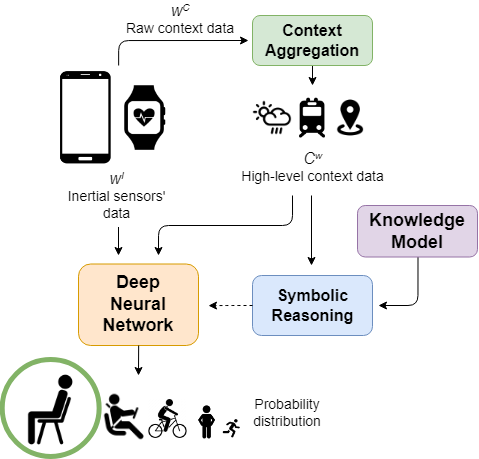}
\caption{The neuro-symbolic context-aware HAR approach}
\label{fig:problem}
\end{figure}

\begin{definition}[Neuro-Symbolic Context-Aware HAR model]
%Given an activity classified based on a deep neural network model $DNN$, a dataset $D^\star$, an annotations set $L$, a knowledge model $K$, and a symbolic reasoner $SR$, the goal of \textit{Neuro-Symbolic Context-Aware Human Activity Recognition} is to learn the weights of $DNN$ and, at the same time, to exploit $K$ and $SR$ to solve the context-aware HAR problem.
A \textit{Neuro-Symbolic Context-Aware Human Activity Recognition} model combines a deep learning model $DNN$ and the symbolic reasoning function $SR()$ to solve the context-aware HAR problem. 
\end{definition}
Figure~\ref{fig:problem} graphically illustrate the high-level structure of NeSy Context-Aware HAR. As we will describe in Sections \ref{sec:revisiting} and \ref{sec:methodology}, the $DNN$ and the $SR()$ modules can be combined in different ways, based on the specific NeSy context-aware HAR technique.

%provide to an unseen tuple $\langle w^I, C^w \rangle$, derived from a sensor data window $w$ from user $u$, the probability distribution $P = \langle p_1, \dots, p_k \rangle$, where $p_i$ is the probability that $u$ performed the activity $a_i$, with $\sum_{i=1}^{k} p_i = 1$. 
%The \textit{Neuro-Symbolic Context-Aware Human Activity Recognition} problem consists 

%To achieve this goal, it is possible to obtain thanks to a dedicated \textsc{context aggregation} module a stream $C_D$ of high-level context data about the surrounding context of a user, starting from the sensors' data streams included in the dataset $D$.  

\subsection{Revisiting Existing Neuro-Symbolic Approaches}
\label{sec:revisiting}

%riformuliamo altri approcci con nostra notazione per appropriata comparazione.

In this section, we re-formulate existing Neuro-Symbolic AI (NeSy) approaches with the notation introduced in sections~\ref{sec:context-aware} and \ref{sec:neuro-symbolic} to compare them with our novel NeSy approach in an appropriate way.
In particular, we consider two state-of-the-art approaches for NeSy HAR: \textit{context refinement} and \textit{symbolic features}.

%Finally, we present a novel methodology for error spotting, that relies on the explainability of neuro-symbolic models.

%Let us consider that the deep neural network $DNN$ is composed of an input layer $l_{inp}$, a sequence of hidden layers $l_{h_1}, \dots, l_{h_m}$, and an output layer $l_{out}$, used for the final classification. The training process of $DNN$ is guided by the categorical cross-entropy loss function $\mathcal{L}_{cross}$, a standard loss function for deep learning architectures. After the training process, $DNN$ periodically emits a probability distribution $P = \langle p_1, \dots, p_k \rangle$ over the possible activities performed by a user, based on the received input data.
%In order to answer the main research question about how to integrate the $SR$ function (described in Section~\ref{sec:problem}) into $DNN$, we propose a novel Neuro Symbolic AI method for HAR called \textit{knowledge infusion through semantic loss} (\textit{semantic loss} for short). We compare our method with two existing solutions named \textit{context refinement} and \textit{knowledge infusion through symbolic features} (or just \textit{symbolic features}), respectively. Specifically, we emphasize how these three approaches differently integrate the \textit{symbolic reasoning} function $SR$ in $DNN$, analyzing pros and cons of each method.

\subsubsection{Context refinement}
\label{sec:context-refinement}

\begin{figure}[h!]
\centering
\includegraphics[width=0.7\linewidth]{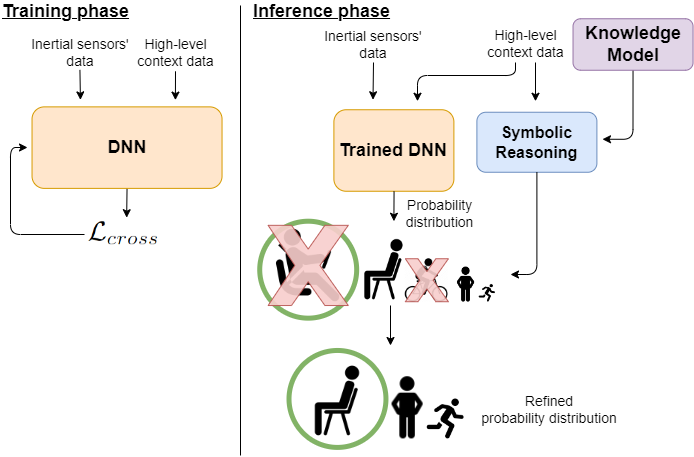}
\caption{The \textit{context refinement} neuro-symbolic approach}
\label{fig:cf_arch}
\end{figure}

The goal of the \textit{context refinement} method~\cite{bettini2020caviar} is to revise a posteriori the $DNN$ predictions through the HAR knowledge encoded into $K$. As shown in Figure \ref{fig:cf_arch}, the $DNN$ is trained with the cross-entropy loss function $\mathcal{L}_{cross}$, which penalizes misclassifications on the training data. During classification, the output of the $SR()$ function is used to exclude from the probability distributions derived by $DNN$ on a specific input those activities that are unlikely to be the correct predictions considering the current context of the user. 

More formally, given a probability distribution $P = \langle p_1, \dots, p_k \rangle$ emitted by $DNN$ on a tuple $\langle w^I, C^w \rangle$, and the set $A^\star$ of context-consistent activities provided by $SR(K,C^w)$, the probability $p_i$ is discarded from $P$ if $a_i \notin A^\star$, and $P$ is normalized to obtain again a probability distribution over the context-consistent activities of $A^\star$. 
%i.e., if the corresponding activity is context-inconsistent according to $SR$. Hence, 

The objective of \textit{context refinement} is to correct a significant number of wrong decisions made by $DNN$, thus increasing its recognition rate. At the same time, it ensures that each classified activity is consistent with the surrounding context of the user. However, computing $SR()$ is strictly required during the inference phase of $DNN$, and its high computational cost complicates the deployment of the \textit{context refinement} method on mobile devices. Moreover, if $K$ encodes rigid constraints about the relationships between contexts and activities, \textit{context refinement} would discard activities that are occasionally performed in unusual context scenarios (e.g., \textit{running} at the mall).

\subsubsection{Symbolic features}
\label{sec:symbolic-features}

\begin{figure}[h!]
\centering
\includegraphics[width=0.7\linewidth]{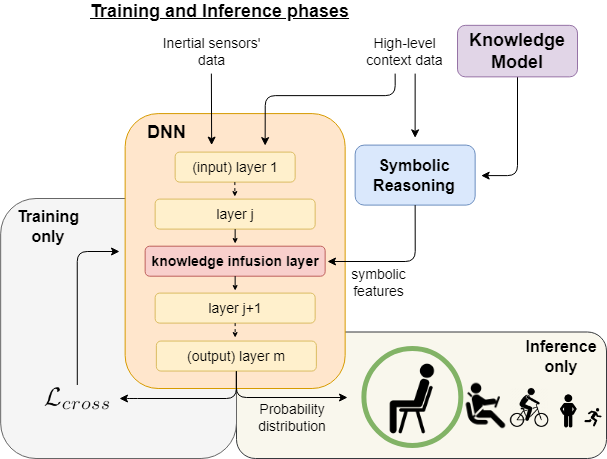}
\caption{The \textit{symbolic features} neuro-symbolic approach}
\label{fig:symbolic_features_arch}
\end{figure}

The objective of the \textit{symbolic features} method~\cite{arrotta2022knowledge} is to directly incorporate the knowledge encoded in $K$ into $DNN$, not only at the inference phase, but also during the learning process. Hence, the \textit{symbolic features} method allows the $DNN$ also to learn the correlations between input data and context-consistent activities. As depicted in Figure~\ref{fig:symbolic_features_arch},
the information about the context-consistent activities provided by $SR$ is used to generate symbolic features that are infused within the hidden layers of $DNN$ through a dedicated layer named \textit{knowledge infusion} layer. More formally, given an input tuple $\langle w^I, C^w \rangle$, and the set $A^\star$ of context-consistent activities provided by $SR(K,C^w)$, the symbolic features consist of a vector $f_s$ in which the $i$-th element is $1$ if $a_i \in A^\star$, $0$ otherwise.
Given the sequence of $DNN$'s layers $\ell_{1}, \dots, \ell_{m}$, and the symbolic features $f_s$ generated through $SR$, the \textit{symbolic features} method adds to $DNN$ a \textit{knowledge infusion} layer $\ell_{ki}$. This layer receives as input the symbolic features $f_s$ and the features automatically extracted by a $DNN$'s hidden layer $\ell_{j}$ with $1 < j < m$. Then, $\ell_{ki}$ concatenates in the latent space the features received as input and generates a novel feature vector that is provided to the next layer $\ell_{j+1}$. Also in this case, the $DNN$ is trained through the cross-entropy loss function $\mathcal{L}_{cross}$.

This methodology is less rigid than \textit{context refinement} since domain knowledge is infused into the data-driven model instead of being used to strictly discard those activities that are context-inconsistent according to $K$. On the other hand, similarly to \textit{context refinement}, the main problem of \textit{symbolic features} is that it is challenging to deploy it on mobile devices since symbolic features must also be inferred during classification.

\section{Knowledge Infusion through Semantic Loss}
\label{sec:methodology}

The existing methods introduced in Section~\ref{sec:revisiting} require running symbolic reasoning during classification. As already discussed in the introduction, it is well-known that such approaches are not suitable for resource-constrained devices like mobile/wearable devices~\cite{bobed2015semantic,bettini2020caviar}.
In this section, we present our novel approach that is named \textit{knowledge infusion through semantic loss} (or \textit{semantic loss} for short). Our method generates an activity classifier encoding knowledge-based constraints without requiring symbolic reasoning during the inference phase. 
Hence, a model based on \textit{semantic loss} can be trained offline on a cloud-based server and then deployed on the users' mobile/wearable devices to locally perform real-time activity recognition efficiently.

\subsection{Methodology}
\begin{figure}[h!]
\centering
\includegraphics[width=0.9\linewidth]{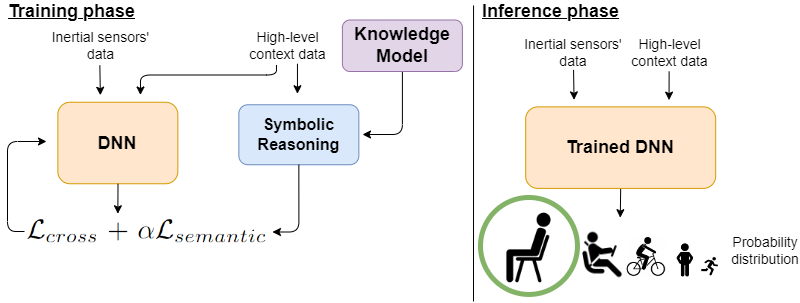}
\caption{Our neuro-symbolic approach based on semantic loss}
\label{fig:semantic_loss_arch}
\end{figure}

In the following, we describe the mechanisms of our \textit{semantic loss} approach.
As depicted in Figure \ref{fig:semantic_loss_arch}, 
the goal of \textit{semantic loss} is to exploit the knowledge $K$ to guide the learning process of $DNN$ through a specifically designed loss function. As in the \textit{symbolic features} method, $DNN$ still learns the correlations between context-consistent activities and input data. At the same time, since no additional features are infused into $DNN$, the use of $K$ and $SR$ during classification is not required, thus solving one of the main limits of the existing solutions. 

Specifically, the loss function $\mathcal{L} = \mathcal{L}_{cross} + \alpha \mathcal{L}_{semantic}$ that guides the training process of $DNN$ is a combination of the cross-entropy loss function $\mathcal{L}_{cross}$ with a semantic loss function $\mathcal{L}_{semantic}$. 
Consistently with other works in the DL literature~\cite{chen2019looks, wu2019deep}, $\alpha$ is a trade-off parameter in charge of balancing the different loss terms. In particular, $\mathcal{L}_{semantic}$ determines how much the $DNN$'s output satisfies the
%common-sense
constraints about the HAR domain encoded into $K$. 
%$\alpha$ is the weight applied to the semantic component of this combined loss.
More formally, given a probability distribution $P = \langle p_1, \dots, p_k \rangle$ emitted by $DNN$ on a tuple $\langle w^I, C^w \rangle$, and the set $A^\star$ of context-consistent activities provided by $SR(K,C^w)$, we denote with $\hat{p} \in P$ the maximum probability value of $P$, and  $\hat{a} \in A$ its corresponding activity.
In the following, we describe five alternative semantic loss functions we designed and tested for this work.
\begin{enumerate}
    \item The \textit{AllConsistentActs (All)} semantic loss focuses on the whole probability distribution $P$ emitted by $DNN$.
    %and not only on the most likely predicted activity $\hat{a}$. 
    Intuitively, given $P$, this semantic loss has the objective of training the network to maximize
    %, for each emitted probability distribution, 
    the sum of the probability values in $P$ that correspond to the context-consistent activities in $A^{\star}$. Hence, we would expect that $DNN$ learns to emit non-zero probabilities only for context-consistent activities during classification. Equation \ref{eq:all} formally defines the \textit{All} semantic loss: 
    \begin{equation}
    \label{eq:all}
        \mathcal{L}_{semanticAll}(P,A^{\star}) = 1 - \sum_{i} p_i \forall i \mid a_i \in A^\star
    \end{equation}
    %More specifically, the penalty applied by the \textit{All} semantic loss is $1 - \sum_{i} p_i$ $\forall i \mid a_i \in A^\star$. 
    A potential drawback of this strategy is that, since it aggregates probability values with a sum, different combinations of these values may lead to the same penalty. Hence, the resulting penalties could be poorly informative for $DNN$ to properly learn knowledge constraints. For this reason, the following alternative semantic losses only focus on the most likely activity $\hat{a}$.
    %The main issue of this semantic loss is that it focuses on each whole probability distribution emitted by the $DNN$. Hence, this approach intuitively requires more training data compared to a semantic loss that analyze only the highest probability of each distribution generated by the $DNN$. For this reason, among the semantic losses we designed, this is the only one that analyzes the whole probability distributions produced by the $DNN$.

    \item The \textit{MinusProb-Prob (-PP)} semantic loss aims at associating low probability values with context-inconsistent activities and higher probability values with context-consistent activities. In particular, context-inconsistent predictions are penalized by their probability value. On the other hand, the penalty of context-consistent activities is inversely proportional to their probability.
    %penalizing the context-inconsistent predictions, weighted by their probability value. At the same time, this approach also penalizes context-consistent predictions so that the lower the probability of a context-consistent activity, the higher will be the applied penalty. The intuition behind this loss is that the $DNN$ should learn to associate the lowest probability values to context-inconsistent predictions, and to be more certain about the context-consistent ones, increasing their probability values. 
    More formally,
    \begin{equation}
        \label{eq:-pp}
        \mathcal{L}_{semantic-PP}(\hat{p},\hat{a},A^{\star}) = 
        \begin{cases}
            1 - \hat{p} & \text{if } \hat{a} \in A^\star
            \\
            \hat{p} & \text{otherwise}
        \end{cases}
    \end{equation}    
    %the \textit{-PP} semantic loss provides a penalty of $1 - \hat{p}$ if $\hat{a} \in A^\star$, $\hat{p}$ otherwise. 
    However, a potential drawback of this strategy is that penalty values for consistent activities with relatively low probability values are similar to penalty values for context-inconsistent activities with relatively high probability values.
    
    %relying on the probability values to penalize both context-consistent and context-inconsistent predictions could make the learning process of the $DNN$ more challenging since it is harder to discriminate a context-consistent and a context-inconsistent prediction.
    
    \item The goal of the \textit{Zero-One (01)} semantic loss is to maximize the differences between penalties of context-consistent and context-inconsistent activities. Specifically,
    \begin{equation}
        \label{eq:01}
        \mathcal{L}_{semantic01}(\hat{p},\hat{a},A^{\star}) = 
        \begin{cases}
            0 & \text{if } \hat{a} \in A^\star
            \\
            1 & \text{otherwise}
        \end{cases}
    \end{equation} 
    %the penalty provided by the \textit{01} semantic loss is $0$ if $\hat{a} \in A^\star$ (i.e., if the prediction is context-consistent), $1$ otherwise. 
    The following strategies are refined versions of the \textit{01} loss.
    %A possible issue of this approach is that it leads the $DNN$ to consider in the same way each context-inconsistent decision, regardless of its probability value. This could complicate the learning process since it is hard to understand the difference between context-inconsistent predictions.
    
    \item The \textit{MinusProb-One (-P1)} semantic loss aims at improving the confidence of $DNN$ on context-consistent predictions. Indeed, while the penalty for context-inconsistent activities is fixed, the penalty for context-consistent activities is inversely proportional to the corresponding probability value. Hence, context-consistent activities with low probabilities values are penalized as well.
    % penalizes inconsistent activities independently from their corresponding probability, so that the $DNN$ privileges consistent activities. 
    More formally,
    \begin{equation}
        \label{eq:-P1}
        \mathcal{L}_{semantic-P1}(\hat{p},\hat{a},A^{\star}) = 
        \begin{cases}
            1 - \hat{p} & \text{if } \hat{a} \in A^\star
            \\
            1 & \text{otherwise}
        \end{cases}
    \end{equation} 
    %the \textit{-P1} loss provides $1$ as penalty when $\hat{a} \notin A^\star$ (i.e., when it is inconsistent). At the same time, the penalty $1 - \hat{p}$ is associated when $\hat{a} \in A^\star$. 
    %Hence, when $\hat{a}$ is consistent, the penalty is inversely proportional to $p$.  
    %The intuition behind this strategy is that  inconsistent activities should be penalized independently from the corresponding probability, since the $DNN$ should privilege consistent activities.
    
    \item Finally, the idea of the \textit{Zero-Prob (0P)} semantic loss is that context-consistent activities should not be penalized, while context-inconsistent activities should be penalized directly proportionally to their associated probability values. Hence, $DNN$ should better learn that the higher the probability values of context-inconsistent activities, the higher the penalty. Therefore, \textit{0P} aims at reducing the probability values on context-inconsistent activities.
   % At the same time, the \textit{0P} semantic loss penalizes context-inconsistent predictions based on their probability value
   % so that the $DNN$ tends to reduce the probability associated with context-inconsistent activities.
    More formally,
    \begin{equation}
        \label{eq:0P}
        \mathcal{L}_{semantic0P}(\hat{p},\hat{a},A^{\star}) = 
        \begin{cases}
            0 & \text{if } \hat{a} \in A^\star
            \\
            \hat{p} & \text{otherwise}
        \end{cases}
    \end{equation} 
    %the provided penalty is $0$ if $\hat{a} \in A^\star$, while it is $\hat{p}$ otherwise. 
    %In this case, the severity of the semantic loss penalty applied to a context-inconsistent prediction is directly related to its probability value.
\end{enumerate}

\section{Experimental evaluation}
\label{sec:experiments}

In this section, we describe the experimental evaluation that we carried out to assess the quality of our method based on semantic loss presented in Section~\ref{sec:methodology}, compared to the state-of-the-art NeSy approaches introduced in Section~\ref{sec:revisiting}. First, we introduce the two datasets that we considered in this work. Then we describe our experimental setup: how we pre-processed the datasets, the models used and the evaluation methodology adopted. Finally, we present the results of our evaluation.

\subsection{Datasets}
\label{sec:datasets}
%In the following, we describe the datasets we consider to empirically evaluate our \textit{semantic loss} method introduced in Section~\ref{sec:methodology}.
The evaluation of context-aware HAR approaches requires datasets including both inertial sensor data and contextual information. However, there are a few publicly available datasets with such characteristics. Existing NeSy approaches for context-aware HAR have been evaluated only on scripted and non-public datasets~\cite{bettini2020caviar}.
In this work, we consider a scripted dataset that we collected in a parallel work and a publicly available in-the-wild dataset, both including sensor and context data.

\subsubsection{DOMINO}
\textit{DOMINO}~\cite{arrotta2023domino} is a HAR dataset
we collected as parallel research in our research lab. \textit{DOMINO} includes several context-dependent activities monitored through mobile devices that collected both inertial sensor data and high-level context data. 

In particular, \textit{DOMINO} includes data from $25$ subjects wearing a smartwatch on their dominant hand's wrist and a smartphone in their pocket. Raw sensor data have been collected from the inertial sensors (accelerometer, gyroscope, and magnetometer) installed on both these mobile devices. At the same time, the dataset also includes high-level context data collected by combining public web services and the smartphone's built-in sensors. The measurements of the barometer and the GPS of the smartphone were discretized to provide information about the users' height and speed variations. Moreover, the dataset incorporates the output of the following web services: (1) \textit{Google's Places API} provided the semantic places closest to the user; from this information, it was also derived the presence of the user in an indoor or an outdoor environment; (2) \textit{OpenWeatherMap} provided current local weather conditions (e.g., sunny), while (3) \textit{Transitland} provided transportation routes and stops close to the user; the combination of this information with location data was used to derive whether the user was following a public transportation route.
%\hl{Reviewer 2: With respect to the dataset used in the evaluation, I find it confusing that some sensors are considered to provide context information and other not. I would rather say that there are different types of sensors that measure different parts of the same situation. The context would be coming from the data of the external services that are used or would be information that could be inferred thanks to the sensor data - Claudio: in sostanza dice che il contesto non deve arrivare da sensori. E' considerata info di piu' alto livello ottenuta con reasoning/astrazione dai sensori oppure da servizi esterni (es. GPS non da contesto, bensi coordinate, dalle quali magari insieme ad altri dati si deriva un contesto (l'utente sta facendo shopping). Non posso dargli torto}

\textit{DOMINO} was acquired in a scripted fashion: the volunteers were asked to perform a sequence of indoor/outdoor activities, but they were not told how to execute them. 
%Hence, the amount of activity instances and their duration is limited. 
Also, the volunteers were monitored by the research staff during data acquisition. As a consequence, the variability of context situations is limited.
%For this reason, \textit{DOMINO} includes two alternative versions of context data: the actual context that surrounded the subjects involved in the data collection campaign and augmented context data that were simulated to cover a wider range of context scenarios. The augmented context data make it possible to evaluate NeSy approaches with more realistic and diverse context situations. 
%For instance, in the actual context data, the \textit{running} activity was never executed indoors, even if it is reasonable to perform it in a gym. Hence, for each activity class, the augmented version of the dataset introduces (based on common-sense knowledge about the HAR domain) simulated context scenarios that partially replaced the actual context data. For instance, considering the \textit{running} activity, in the augmented version of the dataset, it is very common that the user performs it outdoors ($90\%$) while it can also occur indoors ($10\%$). \hl{In order to generate the augmented context data of \textit{DOMINO}, we assumed that the inertial sensor data patterns for an activity do not change that much under different context conditions. We evaluated our semantic loss and the alternative NeSy approaches both on the actual and the augmented versions of \textit{DOMINO}. We will refer to them as \textit{DOMINO Actual} and \textit{DOMINO Augmented}, respectively.}
Overall, \textit{DOMINO} contains almost $9$ hours of labeled data ($\approx 350$ activities instances), including $14$ different types of activities: \textit{walking}, \textit{running}, \textit{standing}, \textit{lying}, \textit{sitting}, \textit{stairs up}, \textit{stairs down}, \textit{elevator up}, \textit{elevator down}, \textit{cycling}, \textit{moving by car}, \textit{sitting on transport}, \textit{standing on transport} and \textit{brushing teeth}.

\subsubsection{ExtraSensory}
\textit{ExtraSensory}~\cite{vaizman2017recognizing} is a public dataset for context and activity recognition. It includes inertial and context data collected in the wild from mobile devices of up to $60$ users. Inertial data were collected through each user's personal smartphone (including both iOS and Android devices) and from a smartwatch provided by the researchers. More specifically, the dataset includes raw data measured by the accelerometer, the gyroscope, and the magnetometer of the smartphone, and raw data collected by the accelerometer of the smartwatch. Besides providing raw sensor data, \textit{ExtraSensory} also provides data as handcrafted feature vectors ($138$ features) extracted from the raw measurements collected through inertial and other smartphone sensors (e.g., microphone, luminosity sensor) in 20-second time windows.

Overall, \textit{ExtraSensory} contains about $300k$ minutes of labeled data, including $51$ different labels self-reported by the users and encoding both high-level context information (e.g., at home, with friends, phone in bag, phone is charging) and performed activities (e.g., sitting, bicycling).  

Since it has been collected in the wild, different research groups in the HAR community used \textit{ExtraSensory} to assess the generalization capabilities of activity recognition frameworks in real-world scenarios \cite{cruciani2020feature, tarafdar2021recognition}. Due to the complexity of the dataset, existing HAR methods evaluated on \textit{ExtraSensory} achieved low recognition rates. For instance, by considering as input the raw inertial measurements provided by the accelerometer and the gyroscope of the smartphones, the CNN-based method proposed in~\cite{cruciani2020feature} reached an average macro f1 score of $\approx0.53$, only considering $4$ target activity classes: \textit{idle} (\textit{lying} or \textit{sitting}), \textit{walking}, \textit{running}, and \textit{cycling}. In another work, by considering the handcrafted features of \textit{ExtraSensory}, an AdaBoost classifier reaches $\approx0.63$ of average macro f1 score on $5$ target activities (i.e., \textit{walking}, \textit{standing}, \textit{sitting}, \textit{exercise}, and \textit{sleeping})~\cite{tarafdar2021recognition}. Hence, this dataset represents a challenging benchmark.

\subsection{Experimental Setup}
\label{sec:experimental_setup}
In the following, we describe our experimental setup. 

\subsubsection{Data pre-processing}
Consistently with existing works proposing NeSy approaches for Context-Aware HAR~\cite{bettini2020caviar}, for both datasets, we segmented sensor data into non-overlapping windows of $k=4$ seconds. In the following, we describe the specific pre-processing steps we adopted for each dataset.

\paragraph{DOMINO}
Considering \textit{DOMINO}, we planned to recognize all the $14$ different available activities, by considering the raw inertial measurements collected by the accelerometer and the gyroscope of the smartphone and the smartwatch. Moreover, in our experiments, we considered $6$ different context information types: the presence of the user in \textit{indoor/outdoor} locations, her \textit{semantic place} (e.g., home, office, gym, bar), her discretized \textit{speed} (i.e., null, low, medium, high), her \textit{proximity to public transportation routes}, her discretized \textit{height variation} (i.e., negative, null, positive), and the \textit{weather conditions} (e.g., sunny, rainy). Table~\ref{tab:domino_samples} shows the number of samples involved during our experiments for each activity class of \textit{DOMINO}.
\begin{table}[]
\footnotesize
\centering
\caption{Number of samples for each activity class in \textit{DOMINO}}
\label{tab:domino_samples}
\begin{tabular}{lc}
\toprule
\multicolumn{1}{c}{\textbf{Activity}} & \textbf{Number of samples} \\ \midrule \midrule
Brushing teeth                        & 163                        \\ \midrule
Cycling                               & 323                        \\ \midrule
Elevator down                         & 171                        \\ \midrule
Elevator up                           & 110                        \\ \midrule
Lying                                 & 387                        \\ \midrule
Moving by car                         & 188                        \\ \midrule
Running                               & 334                        \\ \midrule
Sitting                               & 1764                       \\ \midrule
Sitting on transport                  & 213                        \\ \midrule
Stairs down                           & 266                        \\ \midrule
Stairs up                             & 187                        \\ \midrule
Standing                              & 1875                       \\ \midrule
Standing on transport                 & 297                        \\ \midrule
Walking                               & 1378                       \\ \midrule \midrule
\multicolumn{1}{r}{\textbf{Total}}             & \textbf{7656}                       \\ \bottomrule
\end{tabular}
\end{table}

\paragraph{ExtraSensory}
Considering \textit{ExtraSensory}, we planned to recognize $7$ different activities: \textit{bicycling}, \textit{lying down}, \textit{moving by car}, \textit{on transport}, \textit{sitting}, \textit{standing}, and \textit{walking}. Specifically, for the activity class \textit{walking} we consider those samples labeled as \textit{walking} and/or \textit{strolling} in the original dataset. For \textit{moving by car}, we consider samples labeled with \textit{in a car}, \textit{car driver}, and/or \textit{car passenger}, even when coupled with the label \textit{sitting}. Finally, we labeled as \textit{on transport} those samples originally labeled with \textit{sitting} or \textit{standing} coupled with the label \textit{on a bus}.
%Note that we could have not distinguished \textit{sitting on transport} from \textit{standing on transport} since after our pre-processing steps the latter activity was only performed by a single user. This would have not allowed us to evaluate the different methodologies through the leave-one-out cross-validation technique.

Before conducting our experiments, we performed some steps of data cleaning. First of all, we considered only those samples including inertial measurements recorded from the accelerometer and the gyroscope of the smartphone and from the accelerometer of the smartwatch. Indeed, for some users of \textit{ExtraSensory}, gyroscope data from smartphones are not available. Moreover, not all of the dataset's users wore the smartwatch during data collection. Then, based on the available self-reported labels, we discarded the data collected while the smartphone's user was in a bag, or on a table. 
%In such cases, the smartphone's inertial sensor data would not be useful to classify the activities.
Indeed, we considered only phone positions that have been commonly considered in the literature (i.e., in the pocket and in hand).
Finally, since the labels of \textit{ExtraSensory} were self-reported by the users involved in the data collection, we discarded samples that we considered unreliable, due to the fact that they included self-reported labels not consistent with the recorded data. For instance, we discard segmentation windows including positive speed values but labeled with static physical activities like \textit{lying}. As another example, we discarded those samples simultaneously labeled with \textit{in a car} and \textit{at home}. %\hl{qui ci sarebbero altri esempi, ma mi sembra esagerato elencarli tutti} 
Table \ref{tab:extrasensory_samples} shows the number of samples for each activity class of \textit{ExtraSensory} after data cleaning. Note that, after our data cleaning process, we considered data overall from $31$ subjects.
\begin{table}[]
\footnotesize
\centering
\caption{Number of samples for each activity class in \textit{ExtraSensory}}
\label{tab:extrasensory_samples}
\begin{tabular}{lc}
\toprule
\multicolumn{1}{c}{\textbf{Activity}} & \textbf{Number of samples} \\ \midrule \midrule
Bicycling                             & 2920                       \\ \midrule
Lying down                            & 3055                       \\ \midrule
Moving by car                         & 2150                       \\ \midrule
On transport                          & 610                        \\ \midrule
Sitting                               & 23905                      \\ \midrule
Standing                              & 14280                      \\ \midrule
Walking                               & 11230                      \\ \midrule \midrule
\multicolumn{1}{r}{\textbf{Total}}    & \textbf{58150}             \\ \bottomrule
\end{tabular}
\end{table}

As inertial sensor data, we considered the raw data measured from the accelerometer and the gyroscope of the smartphone and from the accelerometer of the smartwatch. 

%To obtain high-level context data, we consider manipulated versions of a subset of the data included in the dataset that could be automatically derived in a real-world scenario.
Regarding context data, we considered the ones that can be easily derived from sensors of mobile/wearable devices.
For instance, we considered as input context data the information about the user's semantic place (e.g., at the beach) since it could be derived by combining localization data and external web services, but not the position of the user's smartphone (i.e., in the pocket, in hand). In some cases, we discretized available information: for instance, the \textit{speed} values observed thanks to the GPS were discretized into \textit{null/low/medium/high speed}.
%; the same process was applied to the original features concerning the \textit{maximum speed} and the \textit{location diameter}. 
Other high-level context information was obtained by directly considering available data, like \textit{audio level}, \textit{light level}, \textit{screen brightness}, \textit{battery plugged AC/USB}, \textit{battery charging}, \textit{on the phone}, \textit{ringer mode normal/silent/vibrate}, and the time of the day (e.g., \textit{Time 0-6}, \textit{Time 18-24}). Moreover, we relied on the self-reported label \textit{on a bus}, assuming that similar information could be derived by combining GPS data and web services like \textit{Transitland}, as we did in \textit{DOMINO}. Finally, we considered the semantic locations self-reported by the subjects (i.e., \textit{home}, \textit{workplace}, \textit{school}, \textit{gym}, \textit{restaurant}, \textit{shopping}, \textit{bar}, \textit{beach}). As already mentioned, semantic location information can be derived, for instance, by combining location coordinates data with \textit{Google's Places API}.

\subsubsection{$DNN$'s architecture}
The $DNN$ we used for our experiments receives as input three %different data flows 
separate inputs for each segmentation window: a) the smartphone's inertial sensors data, b) the smartwatch's inertial sensors data, and c) the one-hot encoded high-level context data\footnote{Note that, we did not include raw context data as input since it is intuitively easier to learn correlations between activities and high-level context (e.g., semantic place, discretized speed) rather than between activities and raw context (e.g., geographical coordinates, raw speed measurements).}.

Similarly to existing works, we rely on convolutional neural networks to capture spatiotemporal dependencies of sensor data~\cite{zhao2017convolutional, ronao2016human, ha2015multi, yang2015deep}. Even though more sophisticated networks have been proposed in the literature, in this work we use a simple solution to focus on the contribution of knowledge. 
The exact structure of our own CNN model has been determined empirically. Specifically, inertial sensors' data from the smartphone are processed by three \textit{convolutional layers} composed of $32$, $64$, and $96$ filters with a kernel size equal to $24$, $16$, and $8$, respectively. These layers are separated by \textit{max pooling} layers with a pool size of $4$. After the three \textit{convolutional layers}, we add a \textit{global max pooling} layer, followed by a \textit{fully connected} layer that includes $128$ neurons. The smartwatch inertial sensors' data are provided to another component of $DNN$ that presents the same sequence of layers used to automatically extract features from the smartphone's inertial data. The only difference is that, in this case, the three \textit{convolutional layers} present a kernel size of $16$, $8$, and $4$, respectively. Finally, the high-level context data is provided to a single \textit{fully connected} layer composed of $8$ neurons. The features extracted by these three independent flows are then combined thanks to a \textit{concatenation} layer, which is followed by a \textit{dropout} layer with a dropout rate of $0.1$, and a \textit{fully connected} layer with $256$ neurons, useful to extract meaningful correlations between the concatenated features. The last layer of the network is a \textit{softmax} layer that is in charge of providing a probability distribution over the possible activities.

In our experiments, we use this DNN architecture in four different ways:
\begin{itemize}
    \item As a purely data-driven \textit{baseline}, without further modifications
    \item Enhanced with our \textit{semantic loss} (see Section \ref{sec:methodology})
    \item Enhanced by combining in the \textit{concatenation layer} the \textit{symbolic features} and the features automatically extracted from input data (see Section~\ref{sec:symbolic-features})
    \item As the \textit{DNN} module of the \textit{context refinement} approach (see Section~\ref{sec:context-refinement})
\end{itemize}

\subsubsection{Knowledge model and Symbolic reasoning}
\label{sec:semantic_model}
For this work, we extended the knowledge model $K$ proposed in the paper where the NeSy \textit{context refinement} method was introduced~\cite{bettini2020caviar}.
Specifically, this knowledge model is an ontology encoding domain-based relationships between activities and contexts according to common-sense knowledge.
%
%Figure \ref{fig:knowledge_model} shows a small sample of our knowledge model.
%
%\begin{figure}[htbp]
%\centering
%\includegraphics[width=1\linewidth]{imgs/knowledge_model.png}
%\caption{A small component of our knowledge model}
%\label{fig:knowledge_model}
%\end{figure}
%
For instance, \textit{brushing teeth} is defined as an activity that can only occur in indoor environments (e.g., at home) without any height variations. On the other hand, \textit{on transport} is defined as an activity that can only take place while the user is following a public transportation route.
As Symbolic Reasoning function $SR()$, we use the \textit{consistency checking} task of the ontology. In particular, for each activity, we evaluate if it is \textit{consistent} considering the available context data. For instance, according to this ontology, \textit{brushing teeth} is not consistent with a context in which the user is in an outdoor location.
%is moving through different spaces (positive speed). %the semantic location of the user is \textit{park}.

%At the same time, $K$ is also used to quantitatively evaluate the explainability level of each method, as it will be described in Section \ref{sec:explanation_score}.

\subsubsection{Cross-Validation}
We evaluated the approaches presented in Sections~\ref{sec:revisiting} and~\ref{sec:methodology} by adopting the \textit{leave-k-users-out} cross-validation technique. At each fold, $k$ users are used to populate the test set, while the remaining users are used to populate training ($90\%)$ and validation ($10\%)$ sets. 
We also simulated several data scarcity scenarios by downsampling the available training data at each fold (e.g., $1\%$, $50\%$). 

Considering the \textit{DOMINO} dataset, we considered $k=1$ (leave-one-user-out cross-validation). On the other hand, as also done by other works in the literature~\cite{cruciani2020feature}, for the \textit{ExtraSensory} dataset we choose $k=5$. At each iteration, we used the test set to evaluate the recognition rate of the different approaches in terms of the F1 score.

For the sake of robustness, we run each experiment $5$ times, computing the average f1 score and the $95\%$ confidence interval.
%considered the average of the results obtained for each data scarcity experiment by conducting $5$ runs of the cross-validation process. At each of the $5$ runs, we chose a different seed to randomly downsample training data.
Overall, the training process was based on a maximum of $200$ epochs and a batch size of $32$ samples. We considered an \textit{early stopping} strategy, stopping the learning process when the loss computed on the validation set did not improve for $5$ consecutive epochs.

\subsection{Results}
In the following, we show how our \textit{semantic loss} approach outperforms a purely data-driven classifier in terms of recognition rate both in scripted and in-the-wild scenarios. We also compare our method with the Neuro-Symbolic AI (NeSy) approaches presented in Section~\ref{sec:revisiting}. Although our method does not include the (computationally expensive) symbolic reasoning during classification, it often reaches recognition rates that are close (and sometimes better) to the ones of the other approaches, especially considering the more realistic scenarios of \textit{ExtraSensory}.

\subsubsection{Semantic loss types comparison}
Table \ref{tab:semantic_losses_comparison} compares the recognition rates (in terms of overall f1 score) of the five semantic loss functions presented in Section~\ref{sec:methodology} on \textit{DOMINO} and \textit{ExtraSensory}.
\begin{table}[]
\footnotesize
\centering
\caption{Comparison between the Semantic Loss types on the different datasets}
\label{tab:semantic_losses_comparison}
\begin{tabular}{lcc}
\toprule
                                 & \multicolumn{2}{c}{\textbf{\begin{tabular}[c]{@{}c@{}}Dataset\\ (training set percentage)\end{tabular}}}                                         \\
                                 & \begin{tabular}[c]{@{}c@{}}DOMINO\\ ($100\%$)\end{tabular}             & \begin{tabular}[c]{@{}c@{}}ExtraSensory\\ ($10\%$)\end{tabular}         \\ \midrule \midrule
\textbf{Baseline}                & 0.9024                                                                 & 0.5199                                                                  \\ \midrule
\textbf{MinusProb-Prob (-PP)}    & \begin{tabular}[c]{@{}c@{}}0.9139\\ $\alpha = 5$\end{tabular}          & \begin{tabular}[c]{@{}c@{}}0.5402\\ $\alpha = 4$\end{tabular}           \\ \midrule
\textbf{Zero-One (01)}           & \begin{tabular}[c]{@{}c@{}}0.9042\\ $\alpha = 1$\end{tabular}          & \begin{tabular}[c]{@{}c@{}}0.5270\\ $\alpha = 7$\end{tabular}           \\ \midrule
\textbf{Zero-Prob (0P)}          & \begin{tabular}[c]{@{}c@{}}0.9162\\ $\alpha = 3$\end{tabular}          & \begin{tabular}[c]{@{}c@{}}0.5288\\ $\alpha = 9$\end{tabular}           \\ \midrule
\textbf{AllConsistentActs (ALL)} & \begin{tabular}[c]{@{}c@{}}0.9094\\ $\alpha = 1$\end{tabular}          & \textbf{\begin{tabular}[c]{@{}c@{}}0.5872\\ $\alpha = 30$\end{tabular}} \\ \midrule
\textbf{MinusProb-One (-P1)}     & \textbf{\begin{tabular}[c]{@{}c@{}}0.9261\\ $\alpha = 7$\end{tabular}} & \begin{tabular}[c]{@{}c@{}}0.5298\\ $\alpha = 5$\end{tabular}           \\ \bottomrule
\end{tabular}
\end{table}
To better emphasize the differences in the recognition rates, on \textit{ExtraSensory} we decided to show the results obtained by considering a data scarcity scenario in which only $10\%$ of the training data are available.
Indeed, the number of training samples in \textit{DOMINO} is nearly equal to the number contained in only $10\%$ of the training samples in \textit{ExtraSensory}.
Moreover, Table~\ref{tab:semantic_losses_comparison} also includes the best $\alpha$ value for each semantic loss type\footnote{$\alpha$ values have been determined empirically by performing a grid search in the range $[1,30]$} and the results obtained by the purely data-driven \textit{baseline} that is based on a standard classification loss. 

Each semantic loss strategy leads to an improvement in the recognition rates compared to the \textit{baseline}, with \textit{-P1} achieving the best improvements on \textit{DOMINO} ($\approx+2.5\%$) and \textit{All} on \textit{ExtraSensory} ($\approx+6.5\%$). 
%We observed that only \textit{-PP} reaches similar results compared to the \textit{baseline}. Indeed, as expected,  \textit{-PP} tends to apply similar penalty values to both context-consistent and context-inconsistent predictions. At the same time, the improvement obtained by \textit{All} is still limited since considering the whole probability distribution makes learning knowledge constraints harder for the network. 
Before running the experiments, we expected similar results for \textit{01}, \textit{-P1}, and \textit{0P} since all these strategies aim at maximizing the distance in penalties between consistent and not-consistent activities. While this insight is confirmed on \textit{ExtraSensory}, on \textit{DOMINO} the \textit{01} approach proved to be not very effective in improving the recognition rate. On this dataset, we observed that, besides increasing the difference between the penalties applied to context-consistent and context-inconsistent predictions, it is also crucial to consider the probability values emitted by $DNN$, especially in the case of a context-consistent prediction, as proved by the \textit{-P1} semantic loss. Finally, the improvement of the \textit{All} strategy on \textit{DOMINO} is limited, probably because learning knowledge constraints considering the whole probability distribution is unnecessarily too hard on simple scripted scenarios. On the other hand, this strategy significantly outperforms the others in the more realistic settings included in \textit{ExtraSensory}.

\subsubsection{Comparison with other approaches}
\begin{table}[]
\footnotesize
\centering
\caption{\textit{DOMINO}: Results in terms of macro f1 score and $95\%$ confidence interval}
\label{tab:res_domino_actual}
\begin{tabular}{lcccccccccc}
\toprule
                                                                      & \multicolumn{10}{c}{\textbf{Training set percentage}}                                                                                                                                                                                                                                                                                                                                                                                                                                                                                                                                                                                                                                                                                                                                               \\
                                                                      & 10\%                                                                         & 20\%                                                                         & 30\%                                                                         & 40\%                                                                         & 50\%                                                                         & 60\%                                                                         & 70\%                                                                         & 80\%                                                                         & 90\%                                                                         & 100\%                                                        \\ \midrule \midrule
\textbf{Baseline}                                                     & \begin{tabular}[c]{@{}c@{}}0.5946\\ ($\pm0.008$)\end{tabular}                & \begin{tabular}[c]{@{}c@{}}0.7529\\ ($\pm0.010$)\end{tabular}                & \begin{tabular}[c]{@{}c@{}}0.8268\\ ($\pm0.006$)\end{tabular}                & \begin{tabular}[c]{@{}c@{}}0.8556\\ ($\pm0.011$)\end{tabular}                & \begin{tabular}[c]{@{}c@{}}0.8835\\ ($\pm0.011$)\end{tabular}                & \begin{tabular}[c]{@{}c@{}}0.8917\\ ($\pm0.010$)\end{tabular}                & \begin{tabular}[c]{@{}c@{}}0.8915\\ ($\pm0.006$)\end{tabular}                & \begin{tabular}[c]{@{}c@{}}0.9007\\ ($\pm0.007$)\end{tabular}                & \begin{tabular}[c]{@{}c@{}}0.8965\\ ($\pm0.002$)\end{tabular}                & 0.9024                                                       \\ \midrule
\textbf{\begin{tabular}[c]{@{}l@{}}Semantic\\ loss -P1\end{tabular}}      & \begin{tabular}[c]{@{}c@{}}0.6144\\ ($\pm0.024$)\\ $\alpha = 7$\end{tabular} & \begin{tabular}[c]{@{}c@{}}0.7712\\ ($\pm0.004$)\\ $\alpha = 8$\end{tabular} & \begin{tabular}[c]{@{}c@{}}0.8469\\ ($\pm0.002$)\\ $\alpha = 9$\end{tabular} & \begin{tabular}[c]{@{}c@{}}0.8679\\ ($\pm0.010$)\\ $\alpha = 7$\end{tabular} & \begin{tabular}[c]{@{}c@{}}0.8892\\ ($\pm0.006$)\\ $\alpha = 7$\end{tabular} & \begin{tabular}[c]{@{}c@{}}0.8889\\ ($\pm0.007$)\\ $\alpha = 7$\end{tabular} & \begin{tabular}[c]{@{}c@{}}0.9049\\ ($\pm0.006$)\\ $\alpha = 8$\end{tabular} & \begin{tabular}[c]{@{}c@{}}0.8997\\ ($\pm0.003$)\\ $\alpha = 7$\end{tabular} & \begin{tabular}[c]{@{}c@{}}0.9021\\ ($\pm0.008$)\\ $\alpha = 6$\end{tabular} & \begin{tabular}[c]{@{}c@{}}0.9261\\ $\alpha = 7$\end{tabular} \\ \midrule
\textbf{\begin{tabular}[c]{@{}l@{}}Symbolic\\ features\end{tabular}}  & \begin{tabular}[c]{@{}c@{}}0.7268\\ ($\pm0.008$)\end{tabular}                & \begin{tabular}[c]{@{}c@{}}0.8590\\ ($\pm0.012$)\end{tabular}                & \begin{tabular}[c]{@{}c@{}}0.9107\\ ($\pm0.011$)\end{tabular}                & \begin{tabular}[c]{@{}c@{}}0.9152\\ ($\pm0.009$)\end{tabular}                & \begin{tabular}[c]{@{}c@{}}0.9198\\ ($\pm0.011$)\end{tabular}                & \begin{tabular}[c]{@{}c@{}}0.9237\\ ($\pm0.009$)\end{tabular}                & \begin{tabular}[c]{@{}c@{}}0.9265\\ ($\pm0.004$)\end{tabular}                & \begin{tabular}[c]{@{}c@{}}0.9254\\ ($\pm0.008$)\end{tabular}                & \begin{tabular}[c]{@{}c@{}}0.9277\\ ($\pm0.007$)\end{tabular}                & 0.9408                                                       \\ \midrule
\textbf{\begin{tabular}[c]{@{}l@{}}Context\\ refinement\end{tabular}} & \begin{tabular}[c]{@{}c@{}}0.8192\\ ($\pm0.009$)\end{tabular}                & \begin{tabular}[c]{@{}c@{}}0.8811\\ ($\pm0.007$)\end{tabular}                & \begin{tabular}[c]{@{}c@{}}0.9078\\ ($\pm0.009$)\end{tabular}                & \begin{tabular}[c]{@{}c@{}}0.9178\\ ($\pm0.005$)\end{tabular}                & \begin{tabular}[c]{@{}c@{}}0.9281\\ ($\pm0.012$)\end{tabular}                & \begin{tabular}[c]{@{}c@{}}0.9305\\ ($\pm0.006$)\end{tabular}                & \begin{tabular}[c]{@{}c@{}}0.9225\\ ($\pm0.004$)\end{tabular}                & \begin{tabular}[c]{@{}c@{}}0.9274\\ ($\pm0.005$)\end{tabular}                & \begin{tabular}[c]{@{}c@{}}0.9232\\ ($\pm0.002$)\end{tabular}                & 0.9221                                                       \\ \bottomrule
\end{tabular}
\end{table}

\begin{table}[]
\footnotesize
\centering
\caption{\textit{ExtraSensory}: Results in terms of macro f1 score and $95\%$ confidence interval}
\label{tab:res_extrasensory}
\begin{tabular}{lccccccccc}
\toprule
                                                                      & \multicolumn{9}{c}{\textbf{Training set percentage}}                                                                                                                                                                                                                                                                                                                                                                                                                                                                                                                                                                                                                                                                           \\
                                                                      & 1\%                                                                           & 2.5\%                                                                         & 5\%                                                                           & 7.5\%                                                                         & 10\%                                                                          & 25\%                                                                          & 50\%                                                                          & 75\%                                                                          & 100\%                                                          \\ \midrule \midrule
\textbf{Baseline}                                                     & \begin{tabular}[c]{@{}c@{}}0.3127\\ ($\pm0.023$)\end{tabular}                 & \begin{tabular}[c]{@{}c@{}}0.4279\\ ($\pm0.008$)\end{tabular}                 & \begin{tabular}[c]{@{}c@{}}0.4867\\ ($\pm0.013$)\end{tabular}                 & \begin{tabular}[c]{@{}c@{}}0.5167\\ ($\pm0.016$)\end{tabular}                 & \begin{tabular}[c]{@{}c@{}}0.5199\\ ($\pm0.011$)\end{tabular}                 & \begin{tabular}[c]{@{}c@{}}0.5842\\ ($\pm0.016$)\end{tabular}                 & \begin{tabular}[c]{@{}c@{}}0.6096\\ ($\pm0.007$)\end{tabular}                 & \begin{tabular}[c]{@{}c@{}}0.5813\\ ($\pm0.032$)\end{tabular}                 & 0.6053                                                         \\ \midrule
\textbf{\begin{tabular}[c]{@{}l@{}}Semantic\\ loss All\end{tabular}}      & \begin{tabular}[c]{@{}c@{}}0.3366\\ ($\pm0.027$)\\ $\alpha = 29$\end{tabular} & \begin{tabular}[c]{@{}c@{}}0.4895\\ ($\pm0.010$)\\ $\alpha = 30$\end{tabular} & \begin{tabular}[c]{@{}c@{}}0.5256\\ ($\pm0.016$)\\ $\alpha = 26$\end{tabular} & \begin{tabular}[c]{@{}c@{}}0.5650\\ ($\pm0.016$)\\ $\alpha = 26$\end{tabular} & \begin{tabular}[c]{@{}c@{}}0.5872\\ ($\pm0.014$)\\ $\alpha = 30$\end{tabular} & \begin{tabular}[c]{@{}c@{}}0.6331\\ ($\pm0.013$)\\ $\alpha = 29$\end{tabular} & \begin{tabular}[c]{@{}c@{}}0.6323\\ ($\pm0.011$)\\ $\alpha = 18$\end{tabular} & \begin{tabular}[c]{@{}c@{}}0.6131\\ ($\pm0.011$)\\ $\alpha = 16$\end{tabular} & \begin{tabular}[c]{@{}c@{}}0.6244\\ $\alpha = 17$\end{tabular} \\ \midrule
\textbf{\begin{tabular}[c]{@{}l@{}}Symbolic\\ features\end{tabular}}  & \begin{tabular}[c]{@{}c@{}}0.3418\\ ($\pm0.010$)\end{tabular}                 & \begin{tabular}[c]{@{}c@{}}0.4720\\ ($\pm0.016$)\end{tabular}                 & \begin{tabular}[c]{@{}c@{}}0.5877\\ ($\pm0.025$)\end{tabular}                 & \begin{tabular}[c]{@{}c@{}}0.6359\\ ($\pm0.008$)\end{tabular}                 & \begin{tabular}[c]{@{}c@{}}0.6534\\ ($\pm0.012$)\end{tabular}                 & \begin{tabular}[c]{@{}c@{}}0.6404\\ ($\pm0.010$)\end{tabular}                 & \begin{tabular}[c]{@{}c@{}}0.6216\\ ($\pm0.007$)\end{tabular}                 & \begin{tabular}[c]{@{}c@{}}0.6268\\ ($\pm0.007$)\end{tabular}                 & 0.6205                                                         \\ \midrule
\textbf{\begin{tabular}[c]{@{}l@{}}Context\\ refinement\end{tabular}} & \begin{tabular}[c]{@{}c@{}}0.6324\\ ($\pm0.014$)\end{tabular}                 & \begin{tabular}[c]{@{}c@{}}0.6540\\ ($\pm0.003$)\end{tabular}                 & \begin{tabular}[c]{@{}c@{}}0.6797\\ ($\pm0.003$)\end{tabular}                 & \begin{tabular}[c]{@{}c@{}}0.6656\\ ($\pm0.004$)\end{tabular}                 & \begin{tabular}[c]{@{}c@{}}0.6622\\ ($\pm0.007$)\end{tabular}                 & \begin{tabular}[c]{@{}c@{}}0.6483\\ ($\pm0.010$)\end{tabular}                 & \begin{tabular}[c]{@{}c@{}}0.6258\\ ($\pm0.007$)\end{tabular}                 & \begin{tabular}[c]{@{}c@{}}0.6067\\ ($\pm0.023$)\end{tabular}                 & 0.6190                                                         \\ \bottomrule
\end{tabular}
\end{table}

Tables \ref{tab:res_domino_actual} and \ref{tab:res_extrasensory} compare our best \textit{semantic loss} method (i.e., \textit{-P1} on \textit{DOMINO} and \textit{All} on \textit{ExtraSensory}) with: i) the purely data-driven \textit{baseline}, ii) the \textit{symbolic features} strategy, and iii) the \textit{context refinement} strategy. More specifically, we considered different percentages of available training data for each dataset, thus comparing the approaches in different data scarcity scenarios. 
%For the experiments conducted on downsampled versions of the training data, we report the average result obtained on $5$ different runs and, in round brackets, the margin of error at a $95\%$ confidence level. 
Note that, during the experimental evaluation, we empirically determined the optimal $\alpha$ values of the \textit{semantic loss} for each training set percentage. 
%Figure \ref{fig:overall_macro} compares our \textit{semantic loss} strategy with the baseline and the alternative NeSy approaches.
%Moreover, it also makes a comparison with combinations of the considered Neuro-Symbolic AI methods (i.e., \textit{semantic loss with context refinement} and \textit{symbolic features with context refinement}) and with a purely data-driven solution (i.e., \textit{baseline}) that just relies on the $DNN$ to predict the current activity the user is performing. 

Overall, on each dataset, the NeSy approaches outperform the \textit{baseline}, considering almost all the data scarcity scenarios. This result suggests that traditional symbolic AI approaches have the potential of enhancing the predicting capabilities of purely data-driven deep learning models. 

Focusing on the scripted scenarios of \textit{DOMINO} (Table \ref{tab:res_domino_actual}), the improvement of the \textit{semantic loss} is lower than the other approaches, especially considering data scarcity scenarios. For instance, considering $10\%$ of training data, \textit{semantic loss} leads to a recognition rate boost over the \textit{baseline} of $\approx+2\%$ on \textit{DOMINO}. %and of $\approx+4\%$ on \textit{DOMINO Augmented}. 
On the other hand, \textit{symbolic features} and \textit{context refinement} lead to improvements of $\approx+13\%$ and $\approx+22\%$, respectively. These performance differences become progressively smaller while increasing training data availability. Indeed, when all the available training data are considered, both \textit{semantic loss} and \textit{context refinement} outperform the \textit{baseline} by $\approx+2\%$, while \textit{symbolic features} leads to an improvement of $\approx+4\%$.

On the other hand, different insights are observed when focusing on the realistic scenarios of \textit{ExtraSensory} (Table \ref{tab:res_extrasensory}). %Considering low percentages of labeled data, \textit{context refinement} still leads to the best improvements over the baseline, compared to \textit{symbolic features}, and \textit{semantic loss}. 
Indeed, on this dataset, the differences between the three NeSy approaches are smaller. For instance, considering $10\%$ of training data, the recognition rate improvements of \textit{semantic loss}, \textit{symbolic features}, and \textit{context refinement} are $\approx+7\%$, $\approx+13\%$, and $\approx+14\%$, respectively. 

In general, the \textit{semantic loss} achieves improvements that lie between $\approx+2\%$ and $\approx+7\%$, sometimes outperforming the recognition rates of the other NeSy techniques. Indeed, the \textit{semantic loss} outperforms \textit{context refinement} from $50\%$ to $100\%$ of training data, and it also outperforms \textit{symbolic features} on $100\%$ of training data. Overall, \textit{context refinement} is more effective than methods based on knowledge infusion (i.e., \textit{symbolic features} and \textit{semantic loss}) when the availability of labeled data is drastically low. However, when slightly more training data are available (e.g., $25\%$ on \textit{ExtraSensory}), all the NeSy approaches lead to similar improvements. 
%Moreover, in different scenarios, our \textit{semantic loss} reaches similar or even better results than other NeSy methods.

Our results indicate that our \textit{semantic loss} is effective in capturing relationships between high-level context data and activities with respect to learning them directly from the training set by using purely data-driven models. 
This is especially true on the \textit{ExtraSensory} dataset, where the improvement of \textit{semantic loss} compared to the \textit{baseline} is larger. 
Indeed, \textit{DOMINO} covers a significantly lower variability of context situations compared to \textit{ExtraSensory}, and the relationships between context and activities can be captured more easily by the \textit{DNN}. On the other hand, the in-the-wild nature of \textit{ExtraSensory} implies a significantly more complex learning task that can be partially lightened by knowledge reasoning.

Note that, due to the complexity of the dataset, we achieved relatively low recognition rates on \textit{ExtraSensory}  (e.g., the max F1 score is $\approx0.68$). As described in Section \ref{sec:datasets}, our results are in line with other works on the same dataset~\cite{cruciani2020feature,tarafdar2021recognition}.

Since the computational complexity of symbolic reasoning is not adequate for real-world deployment on resource-constrained devices like smartphones and smartwatches, the choice of the optimal solution should consider a trade-off between recognition rate and efficiency. We believe that our \textit{semantic loss} method is a way more promising approach since it still improves the \textit{baseline} while not requiring symbolic reasoning at all after training.

\subsubsection{Activity-level results}
\begin{figure*}[h!]
\centering
\includegraphics[width=0.75\linewidth]{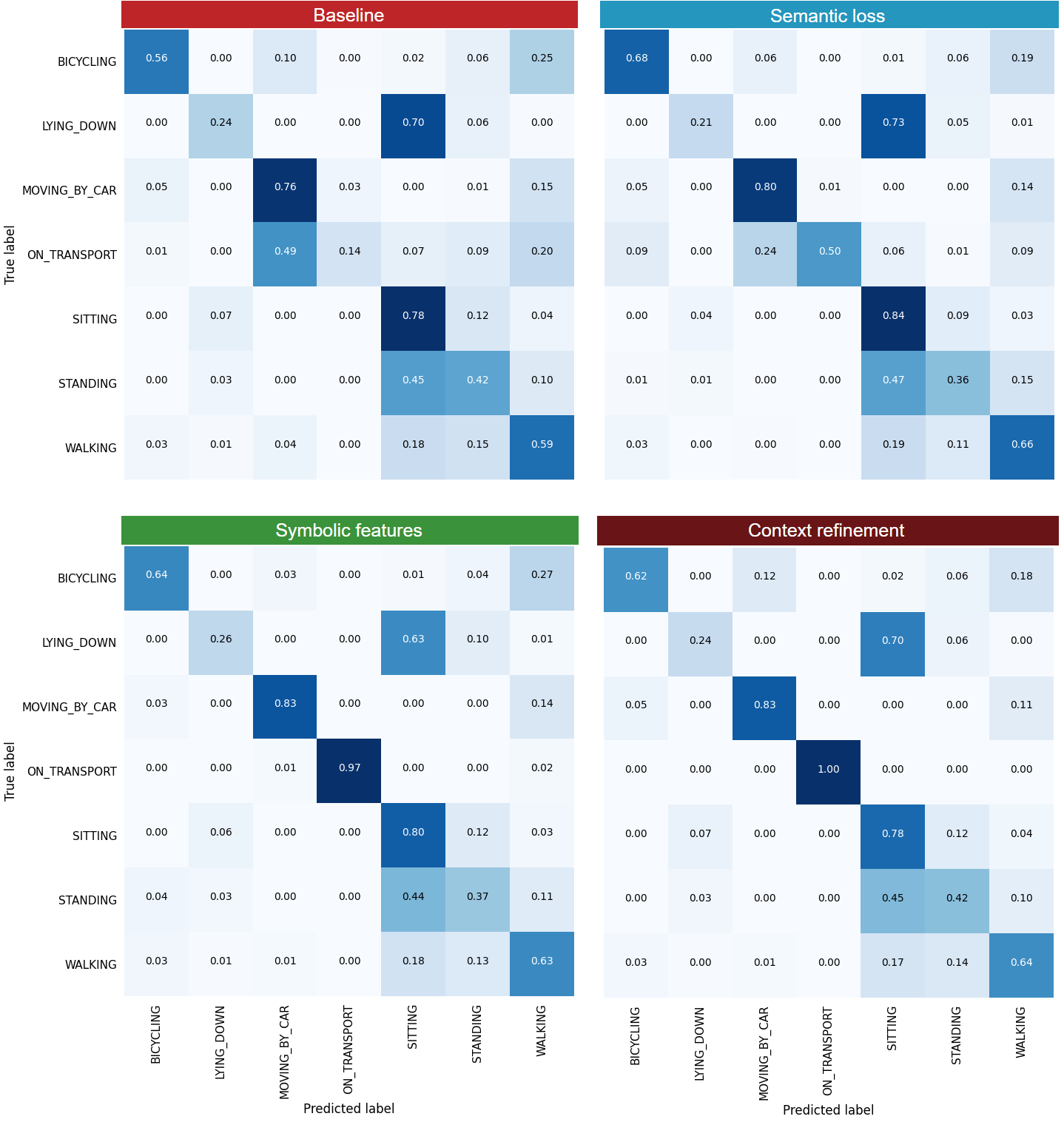}
\caption{Comparison between the confusion matrices of the \textit{baseline} and the three considered Neuro-Symbolic AI approaches trained with $10\%$ of training data on the \textit{ExtraSensory} dataset}
\label{fig:cm_comparison}
\end{figure*}
Figure~\ref{fig:cm_comparison} compares the confusion matrices obtained by the three considered NeSy approaches and the baseline on \textit{ExtraSensory}, considering the data scarcity scenario where only $10\%$ of training data are available\footnote{We show a representative run among the $5$ repetitions of the experiment.}. From these confusion matrices, it emerges the contribution of domain knowledge in improving the recognition of different activities. For instance, the \textit{baseline} often confuses \textit{on transport} with \textit{moving by car} due to their similar patterns (in terms of inertial measurements and speed), even though context information (e.g., whether the user is following a public transportation route) should help in distinguishing them. 

Indeed, even though high-level context data are provided as input to the \textit{baseline}, it is not feasible to learn from the training set all the possible correlations between all the possible context conditions and the performed activities. Hence, enhancing the data-driven model with symbolic AI approaches based on domain knowledge has a key role in enhancing the capabilities of the deep learning model and mitigating this problem, thus significantly reducing the confusion between these two activities.

\section{Strengths and Weaknesses of Neuro-Symbolic Approaches}
This section discusses the strengths and weaknesses of the Neuro-Symbolic AI (NeSy) approaches presented in Sections~\ref{sec:revisiting} and \ref{sec:methodology}. This information is also summarized in Table~\ref{tab:pros_cons}.
\begin{table}[]
\footnotesize
\centering
\caption{Comparison of pros and cons of NeSy methods}
\label{tab:pros_cons}
\begin{tabular}{@{}cccc@{}}

\toprule & 
\begin{tabular}[c]{@{}c@{}}context\\ refinement\end{tabular} & \begin{tabular}[c]{@{}c@{}}symbolic\\ features\end{tabular} & \begin{tabular}[c]{@{}c@{}}semantic\\ loss\end{tabular} \\ \midrule \midrule

\begin{tabular}[c]{@{}c@{}}improving recognition rate\end{tabular}                    & x                                                            & x                                                           & x                                                       \\ \midrule
\begin{tabular}[c]{@{}c@{}}mitigating data scarcity\end{tabular}                      & x                                                            & x                                                           & x                                                       \\ \midrule
\begin{tabular}[c]{@{}c@{}}retraining not required when knowledge changes\end{tabular} & x                                                            &                                                             &                                                         \\ \midrule
\begin{tabular}[c]{@{}c@{}}handling data uncertainty\end{tabular}                     &                                                              & x                                                           & x                                                       \\ \midrule
\begin{tabular}[c]{@{}c@{}}improving DNN's interpretability\end{tabular}         &                                                              & x                                                           & x                                                       \\ \midrule
\begin{tabular}[c]{@{}c@{}}knowledge not required after deployment\end{tabular}        &                                                              &                                                             & x                                                       \\ \bottomrule
\end{tabular}
\end{table}

%\subsection{Comparing the approaches}
Compared to other methods, \textit{context refinement} often reaches the highest recognition rates, especially when the amount of available training data is limited.
Moreover, another advantage of this approach is that it does not require retraining the $DNN$ when the knowledge is updated or changed since symbolic reasoning is applied only during classification. However, this method may be less effective when based on an imperfect knowledge model. Indeed, \textit{context refinement} always discards activities only relying on the user's surrounding context considering rigid constraints. For instance, a user could ride a bicycle even in unusual context scenarios (e.g., on a pedestrian-only road). Hence, when the knowledge model does not cover all the possible contexts in which an activity can be performed, combining the information from inertial data with knowledge would be more convenient in refining the probability distribution.
%Moreover, since \textit{context refinement} is applied after the predictions of the classifier, knowledge-reasoning has no effect in improving the interpretability of the deep decision process that underlies the deep learning model’s prediction.
%
%This allows using the domain knowledge used to generate such additional features for partially explaining the rationale behind each model’s prediction. 
While the \textit{symbolic features} method is sometimes slightly less accurate than \textit{context refinement}, it better captures the intrinsic uncertainty in sensor data, by learning correlations between features and contexts instead of directly applying rigid rules.
%Indeed, the model’s predictions are guided through the symbolic features infused into the hidden layers of the $DNN$. Hence, the knowledge model $K$ can be used to interpret the model's predictions.

However, both approaches require the use of the symbolic reasoning module also after the training phase, complicating their deployment on mobile devices. Indeed, as demonstrated in the literature~\cite{bettini2020caviar, bobed2015semantic}, 
%in a cloud-based architecture, the ontological reasoning process necessary to perform the \textit{context refinement} method requires on average $150$ ms. This process has been performed on a Linux-based machine with an Intel(R) Core(TM) i7-6700 CPU (3.40 GHz), and 16 GB of RAM. On the other hand, the same 
on mobile devices, ontological reasoning could require significantly more than one second of processing time for each window. Since classification is periodically performed every few seconds (e.g., we consider a time window of $4$ seconds), we believe that this computational effort may not be reasonable for real-world deployments.
The main advantage of \textit{semantic loss} is that it
could be trained offline on a server with high computational capabilities and then deployed and used on a mobile device without the need for computationally expensive symbolic reasoning tasks. 
Indeed, \textit{semantic loss} is still able to significantly improve the recognition rate.

Finally, the integration of the domain knowledge into the deep learning model, as performed by our \textit{semantic loss}, has also the potential of making $DNN$ more transparent. Indeed, the decision process of our approach also relies on domain knowledge about HAR that can be used to partially understand from a high-level perspective the overall behavior of the classifier. Clearly, since the knowledge $K$ only considers high-level context data, we are only able to interpret the role of such information during classification. Indeed, inertial sensors' data are not associated with semantics, and explaining their high-level role in the classifier decision process is still an open research problem. However, such interpretability advantages are shared only between the \textit{semantic loss} and the \textit{symbolic features} approaches. Indeed, considering \textit{context refinement}, it is only possible to understand why certain activities are considered consistent or inconsistent. However, the global decision process made by $DNN$ remains opaque since the common-sense knowledge about the specific domain is not directly incorporated into the classifier.

%\subsection{Interpretability aspects}
%
%
%

\section{Conclusion and future work}
In this work, we presented a novel Neuro-Symbolic AI approach for context-aware HAR based on a combination of a classical loss function with a \emph{semantic loss}. Our method infuses domain knowledge inside a deep learning classifier, improving its recognition rate. 
%and the deep model's interpretability. 
Compared to existing neuro-symbolic approaches, our method avoids symbolic reasoning during classification, thus making the model deployment feasible even on devices with limited computational resources. The advantage of our approach is particularly evident in realistic in-the-wild settings.

We have several plans for future work. First, context-Aware HAR requires continuously obtaining context data. However, this may be computationally intensive since it may involve costly operations on mobile/wearable devices (e.g., continuously calling web services). Since high-level contexts may not change so rapidly, we will design strategies to obtain new information periodically (e.g., with a low periodicity, when GPS data exhibits significant changes, etcetera). Thanks to these strategies, it could also be possible to run our method when mobile devices are not connected to the internet for short periods.

Also, our experiments considered a rigid symbolic formalism based on ontologies. However, probabilistic logic frameworks (e.g., Markov Logic Networks) can further improve the results by providing a more flexible knowledge model that considers uncertainty. 

Moreover, we will evaluate how considering knowledge models that encode different levels of detail affect the performance of the NeSy approaches we compared in this work. Indeed, we expect to observe different results when considering a knowledge model that defines only usual contexts in which activities can take place, compared to a knowledge model that instead considers both usual and unusual scenarios.

We also want to explore other strategies to infuse knowledge inside deep learning models. For instance, the symbolic reasoning function may be approximated by a dedicated deep learning model (e.g., through a Graph Neural Network that learns domain constraints from a knowledge graph). This model could significantly reduce symbolic reasoning time, hence making the \textit{context refinement} and \textit{symbolic features} methods more practical in real-world deployments.

Another interesting line of research is to explore whether our approach can be adopted in pervasive computing domains different from HAR, where context information may have a major role (e.g., sensor-based healthcare systems, emotion recognition, anomaly detection).

Finally, we want to study how to quantitatively evaluate the intrinsic interpretability of the DNN components of NeSy approaches.

\bibliographystyle{ACM-Reference-Format}
\bibliography{bibliography}

\end{document}